\documentclass{article} % For LaTeX2e
\usepackage{iclr2017_conference,times}
\usepackage{url}
\usepackage[pdftex]{graphicx}
\usepackage{amsmath,amssymb}
\usepackage{rotating}
\usepackage{colortbl}
\usepackage{placeins}
\usepackage{float}
\usepackage{subfig}
\DeclareMathOperator*{\argmin}{arg\,min}
\DeclareMathOperator{\E}{\mathbb{E}}

\title{Unsupervised Cross-Domain Image Generation}

\author{Yaniv Taigman, Adam Polyak \& Lior Wolf\\
Facebook AI Research\\
Tel-Aviv, Israel \\
\texttt{\{yaniv,adampolyak,wolf\}@fb.com} \\
}

\begin{document}

\maketitle

\begin{abstract}
We study the problem of transferring a sample in one domain to an analog sample in another domain. Given two related domains, $S$ and $T$, we would like to learn a generative function $G$ that maps an input sample from $S$ to the domain $T$, such that the output of a given function $f$, which accepts inputs in either domains, would remain unchanged. Other than the function $f$, the training data is unsupervised and consist of a set of samples from each domain.

The Domain Transfer Network (DTN) we present employs a compound loss function that includes a multiclass GAN loss, an $f$-constancy component, and a regularizing component that encourages $G$ to map samples from $T$ to themselves. We apply our method to visual domains including digits and face images and demonstrate its ability to generate convincing novel images of previously unseen entities, while preserving their identity. 

\end{abstract}

\section{Introduction}

Humans excel in tasks that require making analogies between distinct domains, transferring elements from one domain to another, and using these capabilities in order to blend concepts that originated from multiple source domains. Our experience tells us that these remarkable capabilities are developed with very little, if any, supervision that is given in the form of explicit analogies.

Recent achievements replicate some of these capabilities to some degree: Generative Adversarial Networks (GANs) are able to convincingly generate novel samples that match that of a given training set; style transfer methods are able to alter the visual style of images; domain adaptation methods are able to generalize learned functions to new domains even without labeled samples in the target domain and transfer learning is now commonly used to import existing knowledge and to make learning much more efficient. 

These capabilities, however, do not address the general analogy synthesis problem that we tackle in this work. Namely, given separated but otherwise unlabeled samples from domains $S$ and $T$ and a multivariate function $f$, learn a mapping $G:S\rightarrow T$ such that $f(x) \sim f(G(x))$. 

In order to solve this problem, we make use of deep neural networks of a specific structure in which the function $G$ is a composition of the input function $f$ and a learned function $g$. A compound loss that integrates multiple terms is used. One term is a Generative Adversarial Network (GAN) term that encourages the creation of samples $G(x)$ that are indistinguishable from the training samples of the target domain, regardless of $x\in S$ or $x\in T$. The second loss term enforces that for every $x$ in the source domain training set, $||f(x) - f(G(x))||$ is small. The third loss term is a regularizer that encourages $G$ to be the identity mapping for all $x \in T$.

The type of problems we focus on in our experiments are visual, although our methods are not limited to visual or even to perceptual tasks. Typically, $f$ would be a neural network representation that is taken as the activations of a network that was trained, e.g., by using the cross entropy loss, in order to classify or to capture identity.

As a main application challenge, we tackle the problem of emoji generation for a given facial image. Despite a growing interest in emoji and the hurdle of creating such personal emoji manually, no system has been proposed, to our knowledge, that can solve this problem. Our method is able to produce face emoji that are visually appealing and capture much more of the facial characteristics than the emoji created by well-trained human annotators who use the conventional tools.

\section{Related work}
\label{sec:relatedwork}
As far as we know, the {\it domain transfer} problem we formulate is novel despite being ecological (i.e., appearing naturally in the real-world), widely applicable, and related to cognitive reasoning~\citep{thewaywethink}. In the discussion below, we survey recent GAN work, compare our work to the recent image synthesis work and make links to unsupervised domain adaptation.

GAN~\citep{gan} methods train a generator network $G$ that synthesizes samples from a 
target distribution given noise vectors. $G$ is trained jointly with a discriminator network $D$, which distinguishes between samples generated by $G$ and a training set from the target distribution. The goal of $G$ is to create samples that are classified by $D$ as real samples.

While originally proposed for generating random samples, GANs can be used as a general tool to measure equivalence between distributions. Specifically, the optimization of $D$ corresponds to taking the most discriminative $D$ achievable, which in turn implies that the indistinguishability is true for every $D$. Formally,~\citet{domaingan} linked the GAN loss to the H-divergence between two distributions of~\citet{bendavid}.

The generative architecture that we employ is based on the successful architecture of~\citet{dcgan}. There has recently been a growing concern about the uneven distribution of the samples generated by $G$ -- that they tend to cluster around a set of modes in the target domain~\citep{gantricks}. In general, we do not observe such an effect in our results, due to the requirement to generate samples that satisfy specific $f$-constancy criteria. 

A few contributions (``Conditional GANs'') have employed GANs in order to generate samples from a specific class~\citep{mirza2014conditional}, or even based on a textual description~\citep{RAYLLS16}. When performing such conditioning, one can distinguish between samples that were correctly generated but fail to match the conditional constraint and samples that were not correctly generated. This is modeled as a ternary discriminative function $D$~\citep{RAYLLS16,introspectivegan}.

The recent work by~\citet{deepsim}, has shown promising results for learning to map embeddings to their pre-images, given input-target pairs. Like us, they employ a GAN as well as additional losses in the feature- and the pixel-space. Their method is able to invert the mid-level activations of AlexNet and reconstruct the input image. In contrast, we solve the problem of unsupervised domain transfer and apply the loss terms in different domains: pixel loss in the target domain, and feature loss in the source domain. 

Another class of very promising generative techniques that has recently gained traction is neural style transfer. In these methods, new images are synthesized by minimizing the content loss with respect to one input sample and the style loss with respect to one or more input samples. The content loss is typically the encoding of the image by a network training for an image categorization task, similar to our work. The style loss compares the statistics of the activations in various layers of the neural network. We do not employ style losses in our method. While initially style transfer was obtained by a slow optimization process~\citep{styletransfer}, recently, the emphasis was put on feed-forward methods~\citep{ulyanov16texture,Johnson2016Perceptual}. 

There are many links between style transfer and our work: both are unsupervised and generate a sample under $f$ constancy given an input sample. However, our work is much more general in its scope and does not rely on a predefined family of perceptual losses. Our method can be used in order to perform style transfer, but not the other way around.  Another key difference is that the current style transfer methods are aimed at replicating the style of one or several images, while our work considers a distribution in the target space. In many applications, there is an abundance of unlabeled data in the target domain $T$, which can be modeled accurately in an unsupervised manner.

Given the impressive results of recent style transfer work, in particular for face images, one might get the false impression that emoji are just a different style of drawing faces. By way of analogy, this claim is similar to stating that a Siamese cat is a Labrador in a different style. Emoji differ from facial photographs in both content and style. Style transfer can create visually appealing face images; However, the  properties of the target domain are compromised. 

In the computer vision literature, work has been done to automatically generate sketches from images, see~\citet{6243138} for a survey. These systems are able to emphasize image edges and facial features in a convincing way. However, unlike our method, they require matching pairs of samples, and were not shown to work across two distant domains as in our method. Due to the lack of supervised training data, we did not try to apply such methods to our problems. However, one can assume that if such methods were appropriate for emoji synthesis, automatic face emoji services would be available.

Unsupervised domain adaptation addresses the following problem: given a labeled training set in $S \times Y$, for some target space $Y$, and an unlabeled set of samples from domain $T$, learn a function $h:T \rightarrow Y$~\citep{ICML2012Chen_416,domaingan}. One can solve the sample transfer problem (our problem) using domain adaptation and vice versa. In both cases, the solution is indirect. In order to solve domain adaptation using domain transfer, one would learn a function from $S$ to $Y$ and use it as the input method of the domain transfer algorithm in order to obtain a map from $S$ to $T$\footnote{The function trained this way would be more accurate on $S$ than on $T$. This asymmetry is shared with all experiments done in this work.}. The training samples could then be transferred to $T$ and used to learn a classifier there. 

In the other direction, given the function $f$, one can invert $f$ in the domain $T$ by generating training samples $(f(x),x)$ for $x\in T$ and learn from them a function $h$ from $f(T) = \{f(x) | x\in T\}$ to $T$. Domain adaptation can then be used in order to map $f(S) = \{f(x) | x\in S\}$ to $T$, thus achieving domain transfer.  Based on the work by~\citet{googlesomething}, we expect that $h$, even in the target domain of emoji, will be hard to learn, making this solution hypothetical at this point.

\section{A baseline problem formulation}
\label{sec:baseline}

Given a set $\mathbf{s}$ of unlabeled samples in a source domain $S$ sampled i.i.d according to some distribution ${\cal D}_S$, a set of samples in the target domain $\mathbf{t} \subset T$ sampled i.i.d from distribution ${\cal D}_T$, a function $f$ from the domain $S \cup T$, some metric $d$, and a weight $\alpha$, we wish to learn a function $G:S \rightarrow T$ that minimizes the combined risk $R = R_{\text{GAN}} + \alpha R_{\text{CONST}}$, which is comprised of

\begin{equation}
R_{\text{GAN}} = \max_D \E_{x \sim {\cal D}_S} \log [1-D(G(x))] + \E_{x \sim {\cal D}_T} \log [D(x)], 
\end{equation}

where $D$ is a binary classification function from $T$, $D(x)$ the probability of the class $1$ it assigns for a sample $x \in T$,  and 
\begin{equation}
R_{\text{CONST}} = \E_{x \sim {\cal D}_S} d(f(x),f(G(x)))
\end{equation}

The first term is the adversarial risk, which requires that for every discriminative function $D$, the samples from the target domain would be indistinguishable from the samples generated by $G$ for samples in the source domain. An adversarial risk is not the only option. An alternative term that does not employ GANs would directly compare the distribution ${\cal D}_T$ to the distribution  of $G(x)$ where $x\sim {\cal D}_S$, e.g., by using KL-divergence.

The second term is the $f$-constancy term, which requires that $f$ is invariant under $G$. In practice, we have experimented with multiple forms of $d$ including Mean Squared Error (MSE) and cosine distance, as well as other variants including metric learning losses (hinge) and triplet losses. The performance is mostly unchanged, and we report results using the simplest MSE solution.

Similarly to other GAN formulations, one can minimize the loss associated with the risk $R$ over $G$, while maximizing it over $D$, where $G$ and $D$ are deep neural networks, and the expectations in $R$ are replaced by summations over the corresponding training sets. However, this baseline solution, as we will show experimentally, does not produce desirable results.

\begin{figure}[H]
\centering
\includegraphics[trim=180 150 170 150, clip, width=0.9\linewidth]{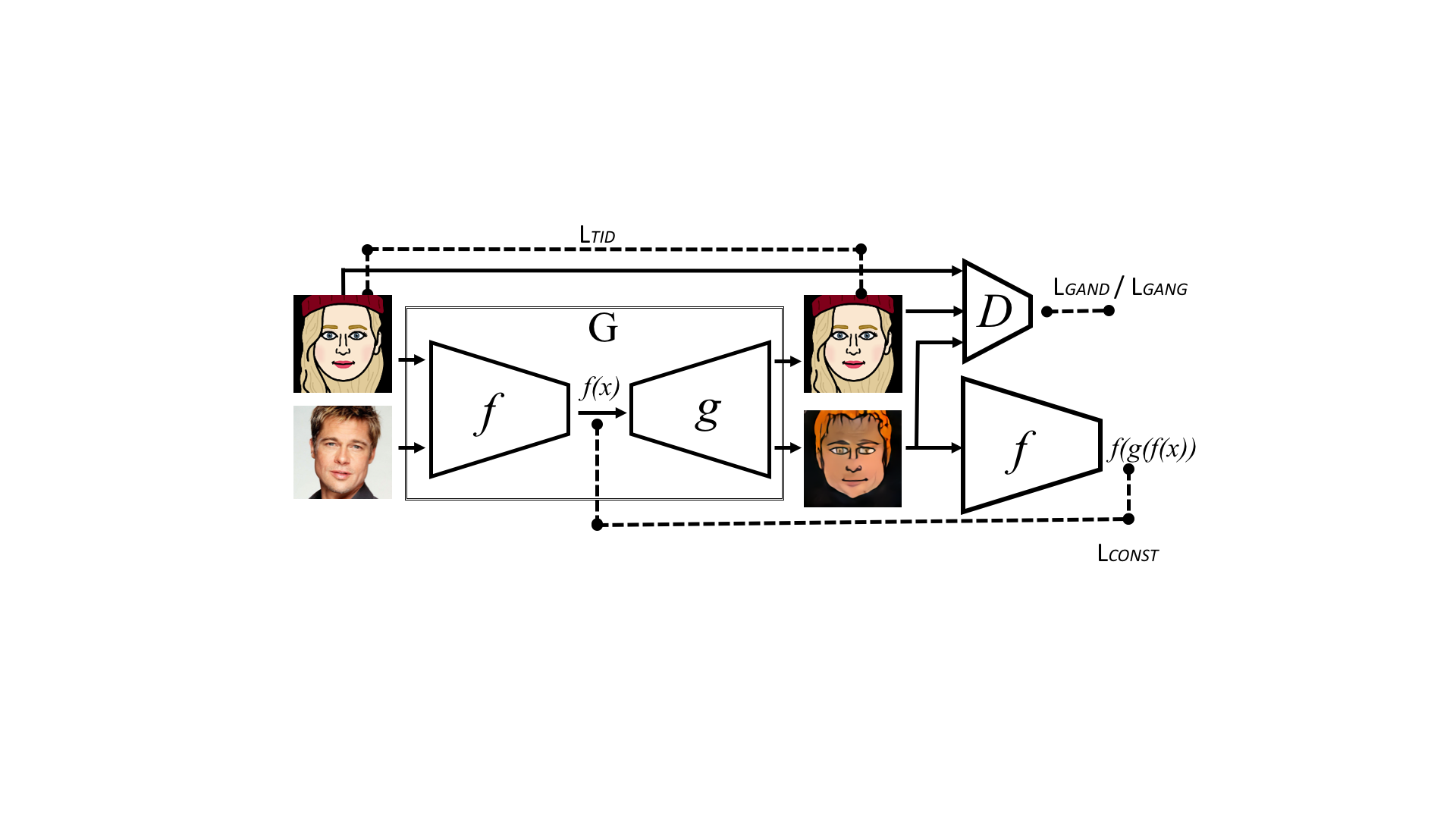}
\caption{\label{fig:illustration}The Domain Transfer Network. Losses are drawn with dashed lines, input/output with solid lines. After training, the forward model G is used for the sample transfer.}
\end{figure}

\section{The domain transfer network}

We suggest to employ a more elaborate architecture that contains two high level modifications. First, we employ $f(x)$ as the baseline representation to the function $G$. Second, we consider, during training, the generated samples $G(x)$ for $x \in \textbf t$. 

The first change is stated as $G=g \circ f$, for some learned function $g$. By applying this, we focus the learning effort of $G$ on the aspects that are most relevant to $R_\text{CONST}$. In addition, in most applications, $f$ is not as accurate on $T$ as it on $S$. The composed function, which is trained on samples from both $S$ and $T$,  adds layers on top of $f$, which adapt it. 

The second change alters the form of $L_\text{GAN}$, making it multiclass instead of binary. It also introduces a new term $L_{TID}$ that requires $G$ to be the identity matrix on samples from $T$. Taken together and written in terms of training loss, we now have two losses $L_D$ and $L_G = L_\text{GANG} + \alpha L_{\text{CONST}} + \beta L_{\text{TID}} + \gamma L_{\text{TV}}$, for some weights $\alpha,\beta,\gamma$, where
\begin{equation}
L_{\text{D}} = -\E_{x \in \mathbf s} \log D_1(g(f(x))) - \E_{x \in \mathbf t} \log D_2(g(f(x))) - \E_{x \in \mathbf t} \log D_3(x) 
\end{equation}
\begin{equation}
L_{\text{GANG}} = -\E_{x \in \mathbf s} \log D_3(g(f(x))) - \E_{x \in \mathbf t} \log D_3(g(f(x)))  
\end{equation}
\begin{equation}
L_{\text{CONST}} = \sum_{x \in \mathbf s} d(f(x),f(g(f(x))))
\end{equation}
\begin{equation}
L_{\text{TID}} = \sum_{x \in \mathbf t} d_2(x,G(x))
\end{equation}

and where $D$ is a ternary classification function from the domain $T$ to ${1,2,3}$, and $D_i(x)$ is the probability it assigns to class $i=1,2,3$ for an input sample $x$, and $d_2$ is a distance function in $T$. During optimization, $L_G$ is minimized over $g$ and $L_D$ is minimized over $D$. See Fig.~\ref{fig:illustration} for an illustration of our method.

The last loss, $L_{\text{TV}}$ is an anisotropic total variation loss~\citep{TV,Mahendran15}, which is added in order to slightly smooth the resulting image. The loss is defined on the generated image $z=[z_{ij}]=G(x)$ as 
\begin{equation} 
L_{TV}(z) =
 \sum_{i,j}
 \left(
 \left(z_{i,j+1} - z_{ij}\right)^2 +
 \left(z_{i+1,j} - z_{ij}\right)^2
 \right)^\frac{B}{2}, 
\end{equation}
where we employ $B = 1$.

In our work, MSE is used for both $d$ and $d_2$. We also experimented with replacing $d_2$, which, in visual domains, compares images, with a second GAN. No noticeable improvement was observed. Throughout the experiments, the adaptive learning rate method Adam by~\citet{adam} is used as the optimization algorithm. 

\begin{figure}[t]
\centering
\subfloat[]{
\includegraphics[trim=0 205 0 0, clip, width=0.1925\linewidth]{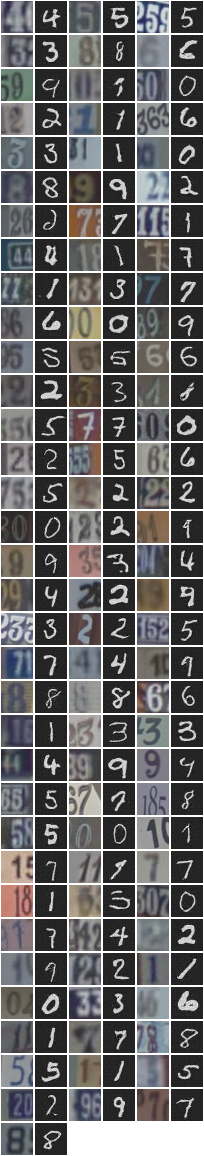}
}
\subfloat[]{
\includegraphics[trim=0 0 0 0, clip, width=0.8\linewidth]{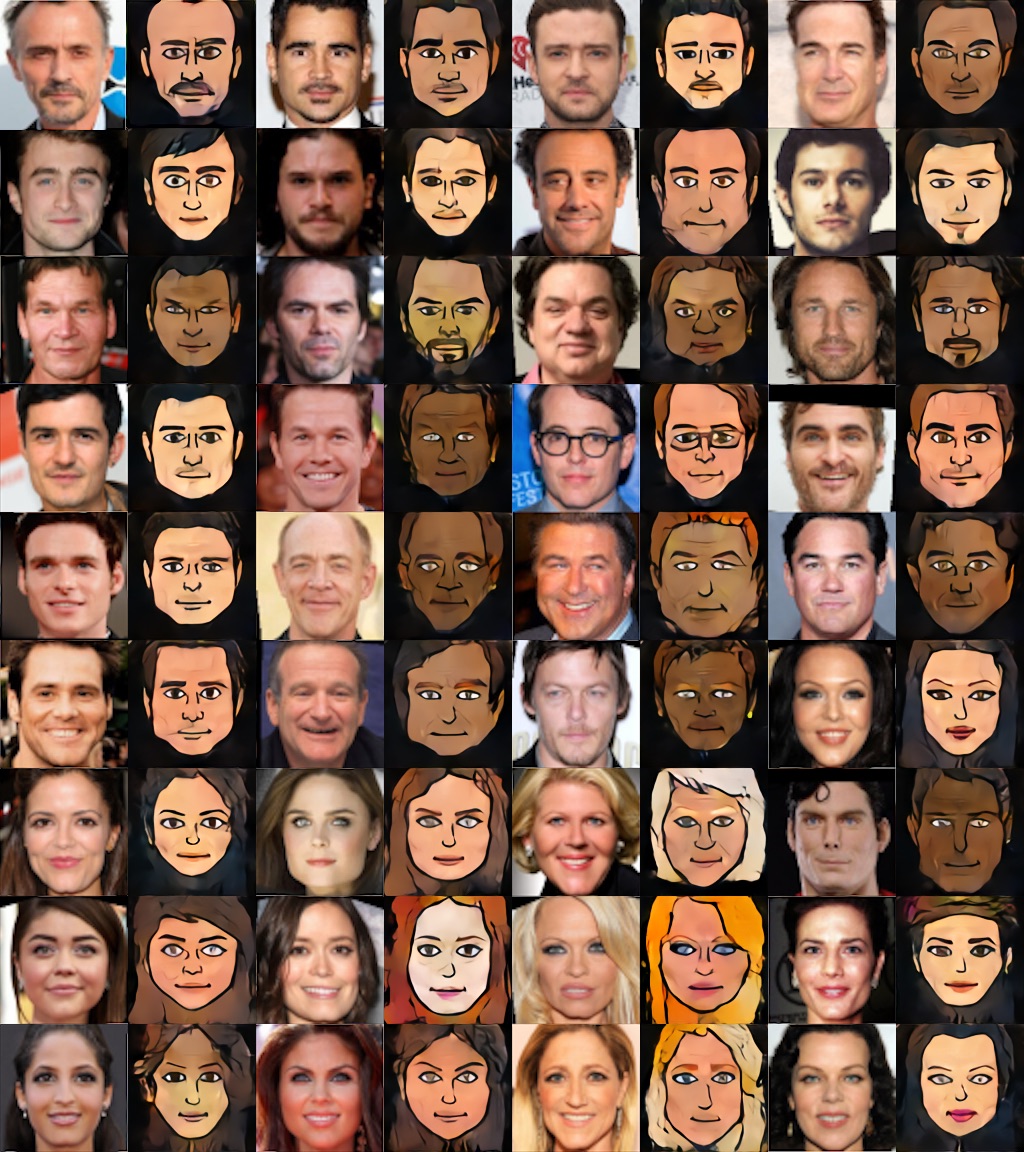}
}
\caption{\label{fig:svnhmnistsamples}Domain transfer in two visual domains. Input in odd columns; output in even columns. (a) Transfer from SVHN to MNIST. (b) Transfer from face photos (Facescrub dataset) to emoji.}
\end{figure}

\section{Experiments}
\label{sec:expa}

The Domain Transfer Network (DTN) is evaluated in two application domains: digits and face images. In the first domain, we transfer images from the Street View House Number (SVHN) dataset of~\cite{svhn} to the domain of the MNIST dataset by~\cite{mnist}. In the face domain, we transfer a set of random and unlabeled face images to a space of emoji images. In both cases, the source and target domains differ considerably.

\subsection{Digits: from SVHN to MNIST}

For working with digits, we employ the extra training split of SVHN, which contains 531,131 images for two purposes: learning the function $f$ and as an unsupervised training set $\textbf s$ for the domain transfer method. The evaluation is done on the test split of SVHN, comprised of 26,032 images. The architecture of $f$ consists of four convolutional layers with ${64, 128, 256,128}$ filters respectively, each followed by max pooling and ReLU non-linearity. The error on the test split is 4.95\%. Even tough this accuracy is far from the best reported results, it seems to be sufficient for the purpose of domain transfer. Within the DTN, $f$ maps a $32 \times 32$ RGB image to the activations of the last convolutional layer of size $128 \times 1 \times 1$ (post a $4 \times 4$ max pooling and before the ReLU). In order to apply $f$ on MNIST images, we replicate the grayscale image three times.

The set $\textbf t$ contains the test set of the MNIST dataset. For supporting quantitative evaluation, we have trained a classifier on the train set of the MNIST dataset, consisting of the same architecture as $f$. The accuracy of this classifier on the test set approaches perfect performance at 99.4\% accuracy, and is, therefore, trustworthy as an evaluation metric. In comparison, the network $f$, achieves 76.08\% accuracy on $\textbf t$. 

Network $g$, inspired by~\cite{dcgan}, maps SVHN-trained $f$'s 128D representations to $32 \times 32$ grayscale images. $g$ employs four blocks of deconvolution, batch-normalization, and ReLU, with a hyperbolic tangent terminal. The architecture of $D$ consists of four batch-normalized convolutional layers and employs ReLU. In the digit experiments, the results were obtained with the tradeoff hyperparamemters $\alpha=\beta=15$. We did not observe a need to add a smoothness term and the weight of $L_\text{TV}$ was set to $\gamma=0$. 

Despite not being very accurate on both domains (and also considerably worse than the SVHN state of the art), we were able to achieve visually appealing domain transfer, as shown in Fig.~\ref{fig:svnhmnistsamples}(a). In order to evaluate the contribution of each of the method's components, we have employed the MNIST network on the set of samples $G(\textbf s_{TEST})=\{G(x)|x\in \mathbf s_{TEST}\}$, using the true SVHN labels of the test set. 

We first compare to the baseline method of Sec.~\ref{sec:baseline}, where the generative function, which works directly with samples in $S$, is composed out of a few additional layers at the bottom of $G$. The results, shown in Tab.~\ref{tab:comparemnist}, demonstrate that DTN has a clear advantage over the baseline method. In addition, the contribution of each one of the terms in the loss function is shown in the table. The regularization term $L_{TID}$ seems less crucial than the constancy term. However, at least one of them is required in order to obtain good performance. The GAN constraints are also important. Finally, the inclusion of $f$ within the generator function $G$ has a dramatic influence on the results. 

\begin{table}[t]
    \begin{minipage}{.5\linewidth}
      \caption{\label{tab:comparemnist} Accuracy of the MNIST classifier on the sampled transferred by our DTN method from SHVN to MNIST.}
      \centering
       \begin{tabular}{lr}
Method & Accuracy \\ \hline \\
Baseline method (Sec.~\ref{sec:baseline})         & 13.71\% \\
\hline
DTN     & {90.66}\%  \\
DTN w/0 $L_\text{TID}$  & 88.40\%  \\
DTN w/0 $L_\text{CONST}$  & 74.55\%  \\
DTN $G$ does not contain $f$  & 36.90\%  \\
DTN w/0 $L_\text{D}$ and $L_\text{GANG}$  & 34.70\% \\
DTN w/0 $L_\text{CONST}$ \& $L_\text{TID}$   &  5.28\%  \\
\hline
Original SHVN image & 40.06\% \\
\hline
\end{tabular}
    \end{minipage}
\hspace{.4cm}
\begin{minipage}{.40\linewidth}
      \centering
        \caption{\label{tab:domaintransfer}Domain adaptation from SVHN to MNIST}
       \begin{tabular}{lr}
Method & Accuracy \\ \hline \\
SA~\citet{fernando} & 59.32\%\\
DANN~\citet{domaingan} & 73.85\%\\
DTN on SVHN transferring\\
the train split $\mathbf s$ & 84.44\%\\
DTN on SVHN transferring\\
the test split & 79.72\%\\
\hline
\end{tabular}
\vspace{.44567847905in}
\end{minipage} 
\end{table}

As explained in Sec.~\ref{sec:relatedwork}, domain transfer can be used in order to perform unsupervised domain adaptation. For this purposes, we transformed the set $\mathbf s$ to the MNIST domain (as above), and using the true labels of $\mathbf s$ employed a simple nearest neighbor classifier there. The choice of classifier was to emphasize the simplicity of the approach; However, the constraints of the unsupervised domain transfer problem would be respected for any classifier trained on $G(\textbf s)$. The results of this experiment are reported in Tab.~\ref{tab:domaintransfer}, which shows a clear advantage over the state of the art method of~\cite{domaingan}. This is true both when transferring the samples of the set $\mathbf s$ and when transferring the test set of SVHN, which is much smaller and was not seen during the training of the DTN.

\subsubsection{Unseen digits}

\begin{figure}[H]
 \centering
 \begin{tabular}{c@{~}c@{~}c@{~}c@{~}c@{~}c}
 \includegraphics[width=0.1597\linewidth]{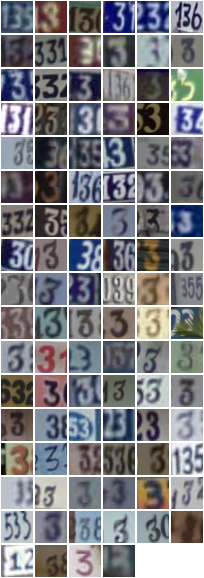}&
 \includegraphics[width=0.1597\linewidth]{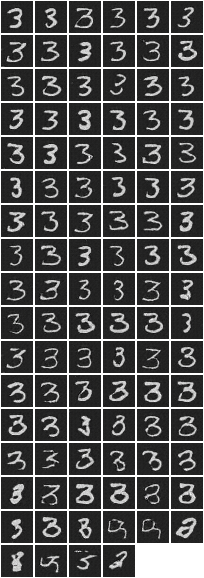}&
\includegraphics[width=0.1597\linewidth]{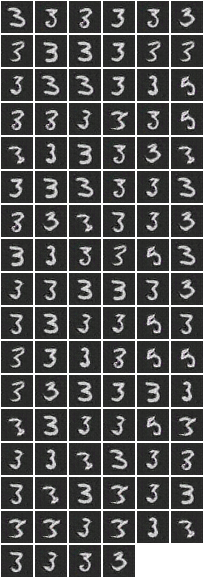}&
 \includegraphics[width=0.1597\linewidth]{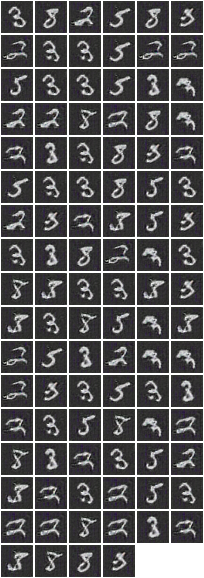}&
\includegraphics[width=0.1597\linewidth]{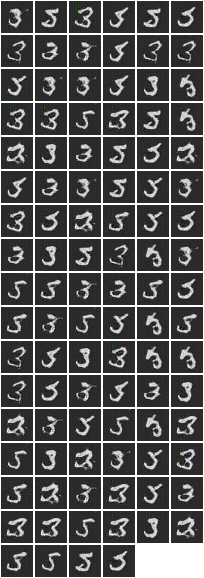}&
 \includegraphics[width=0.1597\linewidth]{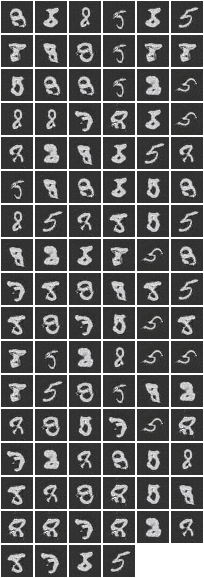}\\
 (a)&(b)&(c)&(d)&(e)&(f)
 \end{tabular}
 \caption{\label{fig:drop3} A random subset of the digit '3' from SVHN, transferred to MNIST. (a) The input images. (b) Results of our DTN. In all plots, the cases keep their respective locations, and are sorted by the probability of `3' as inferred by the MNIST classifier on the results of our DTN. (c) The obtained results, in which the digit 3 was not shown as part of the set $\mathbf s$ unlabeled samples from SVNH.  (d) The obtained results, in which the digit 3 was not shown as part of the set $\mathbf t$ of unlabeled samples in MNIST. (e) The digit 3 was not shown in both $\mathbf s$ and $\mathbf t$. (f) The digit 3 was not shown in $\mathbf s$, $\mathbf t$, and during the training of $f$.}
\end{figure}

Another set of experiments was performed in order to study the ability of the domain transfer network to overcome the omission of a class of samples. This type of ablation can occur in the source or the target domain, or during the training of $f$ and can help us understand the importance of each of these inputs. The results are shown visually in Fig.~\ref{fig:drop3}, and qualitatively in Tab.~\ref{tab:compare3}, based on the accuracy of the MNIST classifier only on the transferred samples from the test set of SVHN that belong to class `3'. 

It is evident that not including the class in the source domain is much less detrimental than eliminating it from the target domain. This is the desirable behavior: never seeing any `3'-like shapes in $\mathbf t$, the generator should not generate such samples. Results are better when not observing `3' in both  $\mathbf s,\mathbf t$ than when not seeing it only in $\mathbf t$ since in the latter case, $G$ learns to map source samples of `3' to target images of other classes.

\begin{table}[t]
\caption{Comparison of recognition accuracy of the digit 3 as generated in MNIST}
\label{tab:compare3}
\begin{center}
\begin{tabular}{lr}
Method & Accuracy of `3'\\ \hline \\
DTN & 94.67\%   \\
`3' was not shown in $\mathbf s$            & 93.33\%  \\
`3' was not shown in $\mathbf t$             &  40.13\% \\
`3' was not shown in both $\mathbf s$ or $\mathbf t$     & 60.02\%   \\
`3' was not shown in $\mathbf s$, $\mathbf t$, and during the training of $f$ & 4.52 \%  \\
\hline
\end{tabular}
\end{center}
\end{table}

\subsection{Faces: from Photos to Emoji}

For face images, we use a set $\mathbf s$ of one million random images without identity information. The set $\mathbf t$ consists of assorted facial avatars ({\it emoji}) created by an online service (\url{bitmoji.com}). The emoji images were processed by a fully automatic process that localizes, based on as set of heuristics, the center of the irides and the tip of the nose. Based on these coordinates, the emoji were centered and scaled into $152 \times 152$ RGB images.

As the function $f$, we employ the representation layer of the DeepFace network~\cite{deepface}. This representation is 256-dimensional and was trained on a labeled set of four million images that does not intersect the set $\mathbf s$. Network D takes $152 \times 152$ RGB images (either natural or scaled-up emoji) and consists of 6 blocks, each containing a convolution with stride 2, batch normalization, and a leaky ReLU with a parameter of 0.2. Network $g$ maps $f$'s 256D representations to $64 \times 64$ RGB images through a network with 5 blocks, each consisting of an upscaling convolution, batch-normalization and ReLU. Adding $1 \times 1$ convolution to each block resulted in lower $L_\text{CONST}$ training errors, and made $g$ 9-layers deep. We set $\alpha = 100$, $\beta = 1$, $\gamma = 0.05$ as the tradeoff hyperparameters within $L_G$ via validation. As expected, higher values of $\alpha$ resulted in better $f$-constancy, however introduced artifacts such as general noise or distortions. 

In order to upscale the $64 \times 64$ output to print quality, we used the method of~\cite{superres}, which was shown to work well on art. We did not retrain this network for our application, and used the published one. Results without this upscale are shown, for comparison, in Appendix~\ref{sec:superres}. 

\begin{figure}[t]
\centering
\begin{tabular}{cccc}
\includegraphics[trim=0 0 0 0, clip, width=0.2235\linewidth]{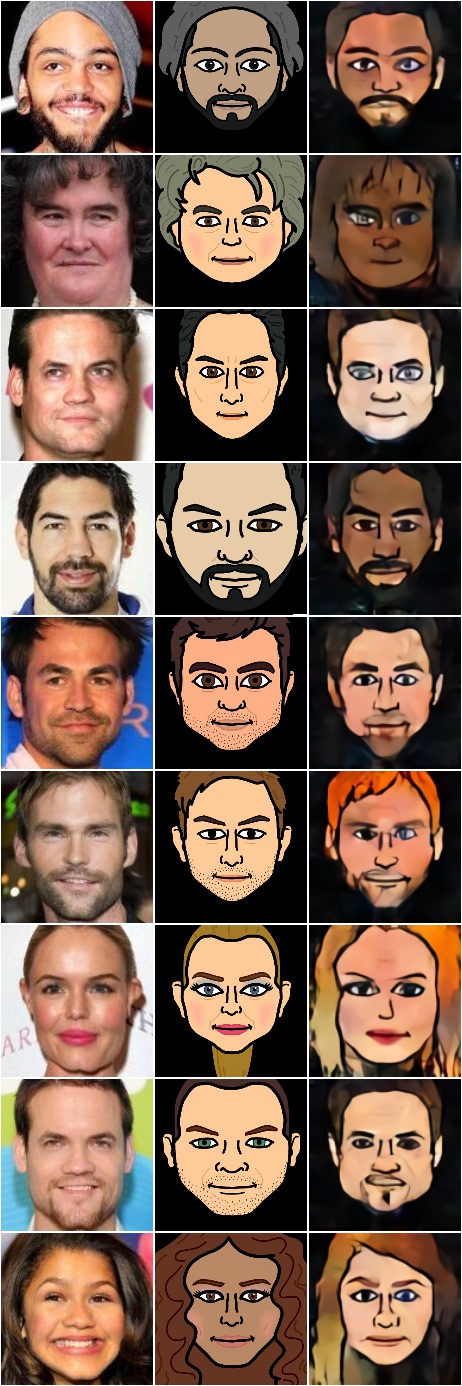}&
\includegraphics[trim=0 0 0 0, clip, width=0.2235\linewidth]{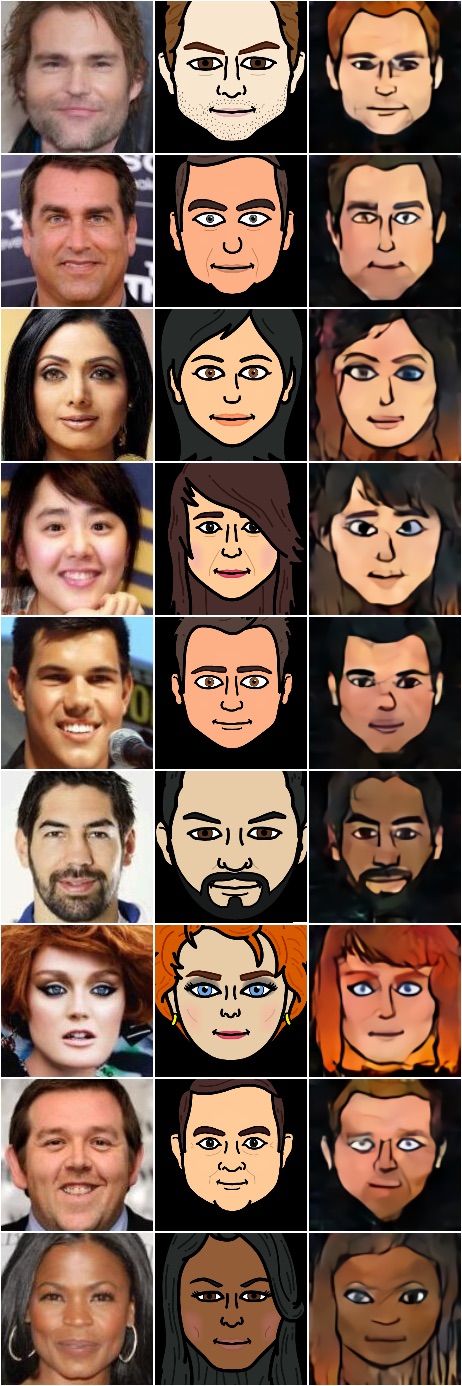}&
\includegraphics[trim=0 0 0 0, clip, width=0.2235\linewidth]{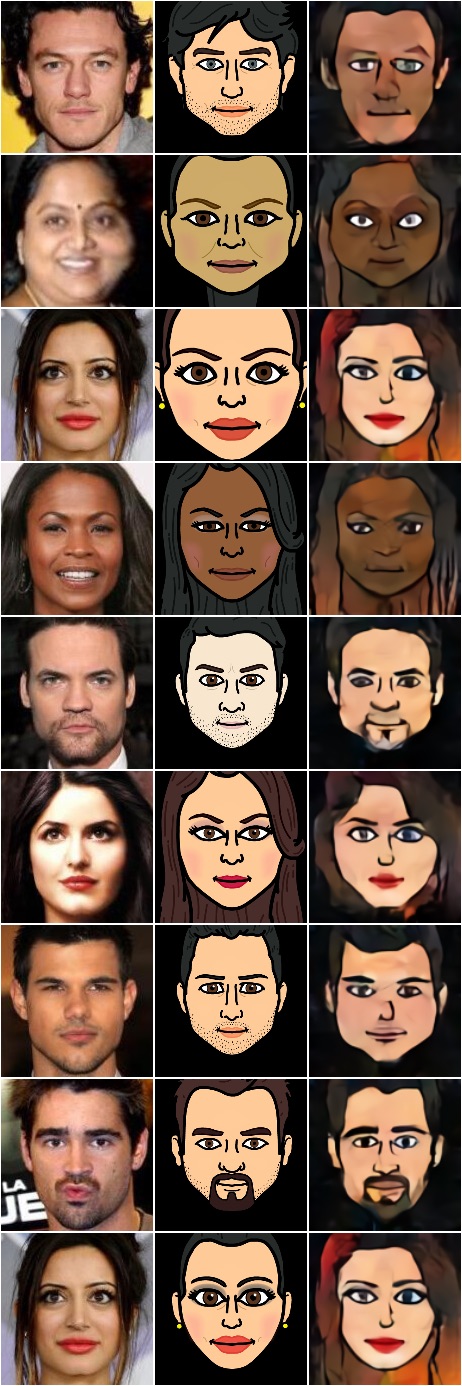}&
\includegraphics[trim=0 0 0 0, clip, width=0.2235\linewidth]{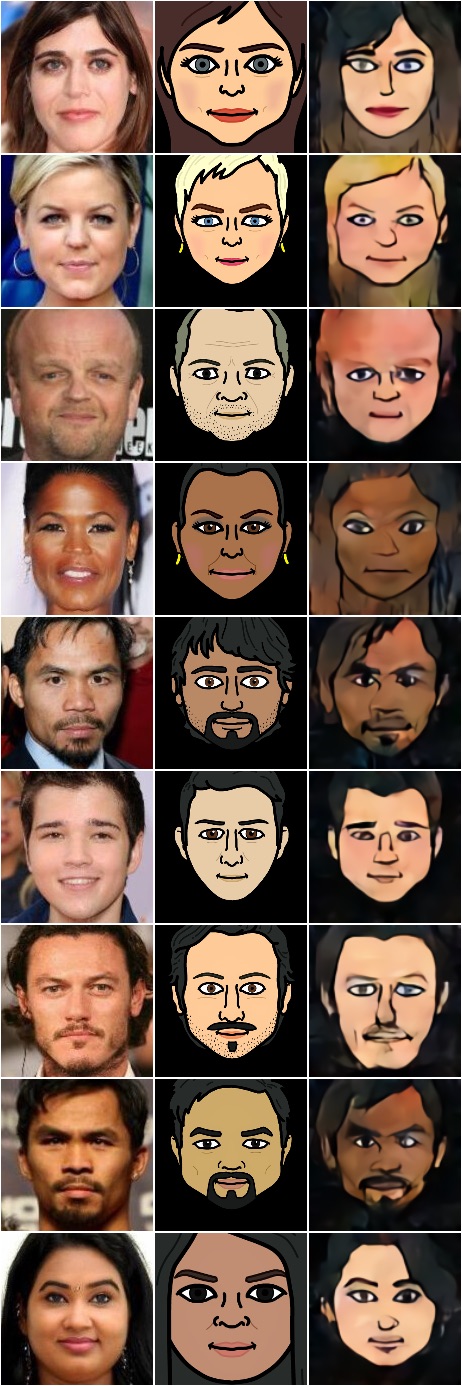}
\end{tabular}
\caption{\label{fig:IBtagB} Shown, side by side are sample images from the CelebA dataset, the emoji images created manually using a web interface (for validation only), and the result of the unsupervised DTN. See Tab.~\ref{tab:BvsG} for retrieval performance.}
\end{figure}

\paragraph{Comparison With Human Annotators}For evaluation purposes only, a team of professional annotators manually created an emoji, using the web service of \url{bitmoji.com}, for $118$ random images from the CelebA dataset~\citep{celeba}.  Fig.~\ref{fig:IBtagB} shows side by side samples of the original image, the human generated emoji and the emoji generated by the learned generator function $G$.  As can be seen, the automatically generated emoji tend to be more informative, albeit less restrictive than the ones created manually. 

In order to evaluate the identifiability of the resulting emoji, we have collected a second example for each identity in the set of $118$ CelebA images and a set $\mathbf s'$ of 100,000 random face images, which were not included in $\mathbf s$. We then employed the VGG face CNN descriptor of~\cite{Parkhi15} in order to perform retrieval as follows. For each image $x$ in our manually annotated set, we create a gallery  $\mathbf s' \cup x'$, where $x'$ is the other image of the person in $x$. We then perform retrieval using the VGG face descriptor using either the manually created emoji or $G(x)$ as probe. 

The VGG network is used in order to avoid a bias that might be caused by using $f$ both for training the DTN and for evaluation. The results are reported in Tab.~\ref{tab:BvsG}. As can be seen, the emoji generated by $G$ are much more discriminative than the emoji created manually and obtain a median rank of 16 in cross-domain identification out of $10^5$ distractors.

\begin{table}[t]
\caption{Comparison of retrieval accuracy out of a set of 100,001 face images for either manually created emoji or the one created by the DTS network.}
\label{tab:BvsG}
\begin{center}
\begin{tabular}{lrr}
Measure & Manual & Emoji by DTN\\ \hline \\
Median rank & 16311 & 16\\
Mean rank & 27,992.34 & 535.47\\
Rank-1 accuracy & 0\% & 22.88\% \\
Rank-5 accuracy & 0\% & 34.75\% \\
\hline
\end{tabular}
\end{center}
\end{table}

\paragraph{Multiple Images Per Person} We evaluate the visual quality that is obtained per person and not just per image, by testing DTN on the Facescrub dataset~\citep{facescrub}. For each person $p$, we considered the set of their images $X_p$, and  selected the emoji that was most similar to their source image:
\begin{equation}
\label{eq:mmeq}
\argmin_{x \in X_p} || f(x)-f(G(x))|| 
\end{equation}

This simple heuristic seems to work well in practice; The general problem of mapping a set $X \subset S$ to a single output in $T$ is left for future work.  Fig.~\ref{fig:svnhmnistsamples}(b) contains several examples from the Facescrub dataset. For the complete set of identities, see Appendix~\ref{sec:facescrub}.

\paragraph{Network Visualization} The obtained mapping $g$ can serve as a visualization tool for studying the properties of the face representation. This is studied in Appendix~\ref{sec:e10} by computing the emoji generated for the standard basis of $\mathbb{R}^{256}$. The resulting images present a large amount of variability, indicating that $g$ does not present a significant mode effect.

\begin{figure}[t]
\centering
\subfloat[]{
\includegraphics[trim=130 0 60 0, clip, width=0.17\linewidth]{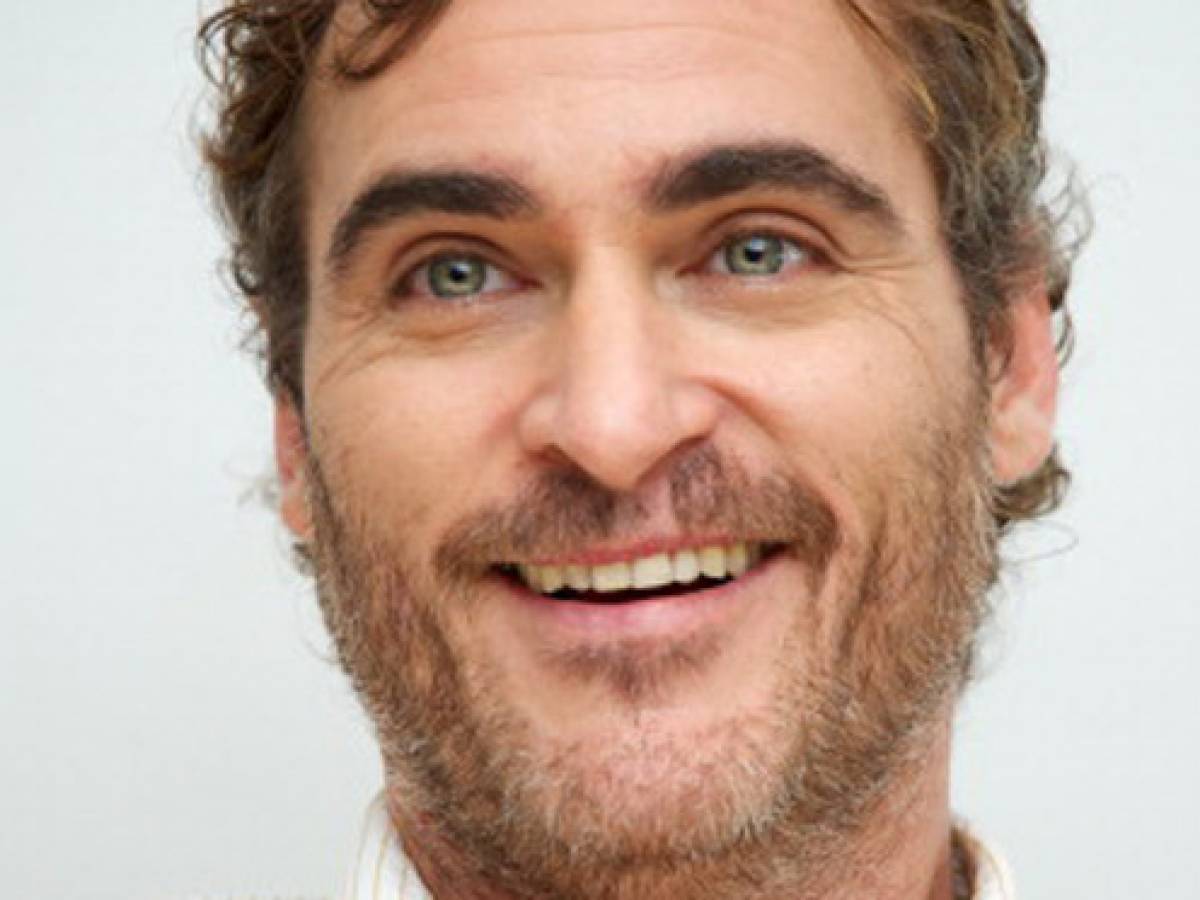}
}
\subfloat[]{
\includegraphics[trim=0 0 0 0, clip, width=0.15\linewidth]{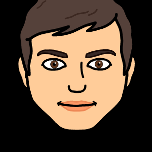}
}
\subfloat[]{
\includegraphics[trim=33 0 16 0, clip, width=0.16\linewidth]{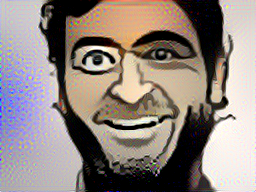}
}
\subfloat[]{
\includegraphics[trim=0 0 0 0, clip, width=0.15\linewidth]{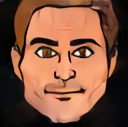}
}
\subfloat[]{
\includegraphics[trim=0 0 0 0, clip, width=0.15\linewidth]{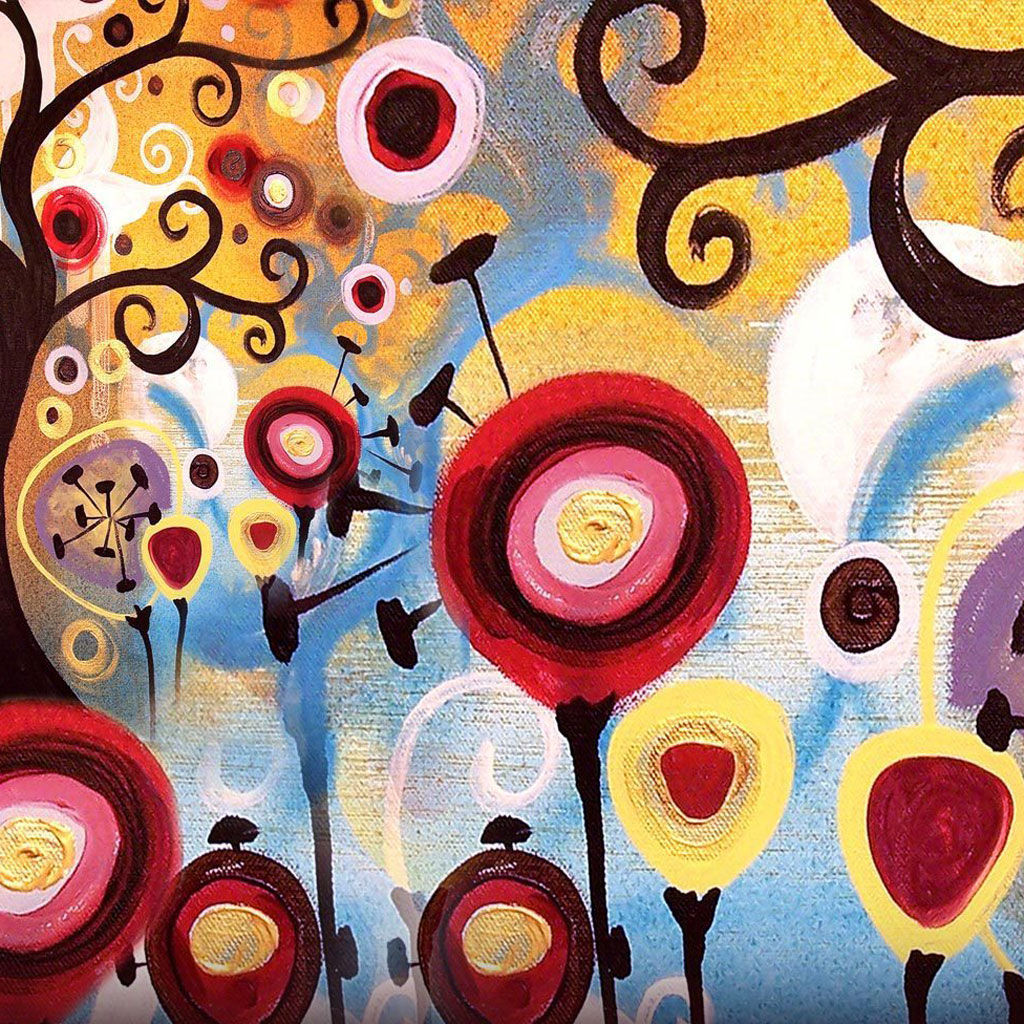}
}
\subfloat[]{
\includegraphics[trim=0 0 0 0, clip, width=0.15\linewidth]{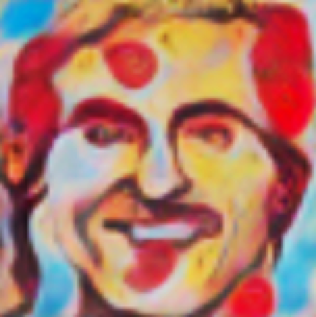}
}

\caption{\label{fig:styletransfer}Style transfer as a specific case of Domain Transfer. (a) The input content photo. (b) An emoji taken as the input style image. (c) The result of applying the style transfer method of~\cite{styletransfer}.  (d) The result of the emoji DTN.  (e)  Source image for style transfer. (f) The result, on the same input image, of a DTN trained to perform style transfer.}
\end{figure}

\subsection{Style Transfer as a specific Domain Transfer task}

Fig.~\ref{fig:styletransfer}(a-c) demonstrates that neural style transfer~\cite{styletransfer} cannot solve the photo to emoji transfer task in a convincing way. The output image is perhaps visually appealing; However, it does not belong to the space $\mathbf t$ of emoji. Our result are given in Fig.~\ref{fig:styletransfer}(d) for comparison. Note that DTN is able to fix the missing hair in the image.

Domain transfer is more general than style transfer in the sense that we can perform style transfer using a DTN. In order to show this, we have transformed, using the method of~\cite{Johnson2016Perceptual}, the training images of CelebA based on the style of a single image (shown in Fig.~\ref{fig:styletransfer}(e)). The original photos were used as the set $\mathbf s$, and the transformed images were used as $\mathbf t$. Applying DTN, using face representation $f$, we obtained styled face images such as the one shown in the figure~\ref{fig:styletransfer}(f).

\section{Discussion and Limitations}
\label{sec:conclusions}
Asymmetry is central to our work. Not only does our solution handle the two domains $S$ and $T$ differently, the function $f$ is unlikely to be equally effective in both domains since in most practical cases, $f$ would be trained on samples from one domain. While an explicit domain adaptation step can be added in order to make $f$ more effective on the second domain, we found it to be unnecessary. Adaptation of $f$ occurs implicitly due to the application of $D$ downstream. 

Using the same function $f$, we can replace the roles of the two domains, $S$ and $T$. For example, we can synthesize an SVHN image that resembles a given MNIST image, or synthesize a face that matches an emoji. As expected, this yields less appealing results due to the asymmetric nature of $f$ and the lower information content in these new source domains, see Appendix~\ref{sec:humanizer}.

Domain transfer, as an unsupervised method, could prove useful across a wide variety of computational tasks. Here, we demonstrate the ability to use domain transfer in order to perform unsupervised domain adaptation. While this is currently only shown in a single experiment, the simplicity of performing domain adaptation and the fact that state of the art results were obtained effortlessly with a simple nearest neighbor classifier suggest it to be a promising direction for future research.

\bibliography{gans}
\bibliographystyle{iclr2017_conference}

\FloatBarrier

\newpage
\appendix

\section{Facescrub dataset generations}
\label{sec:facescrub}
\begin{figure}[H]
\centering
\includegraphics[trim=0 0 0 0, clip, width=1\linewidth]{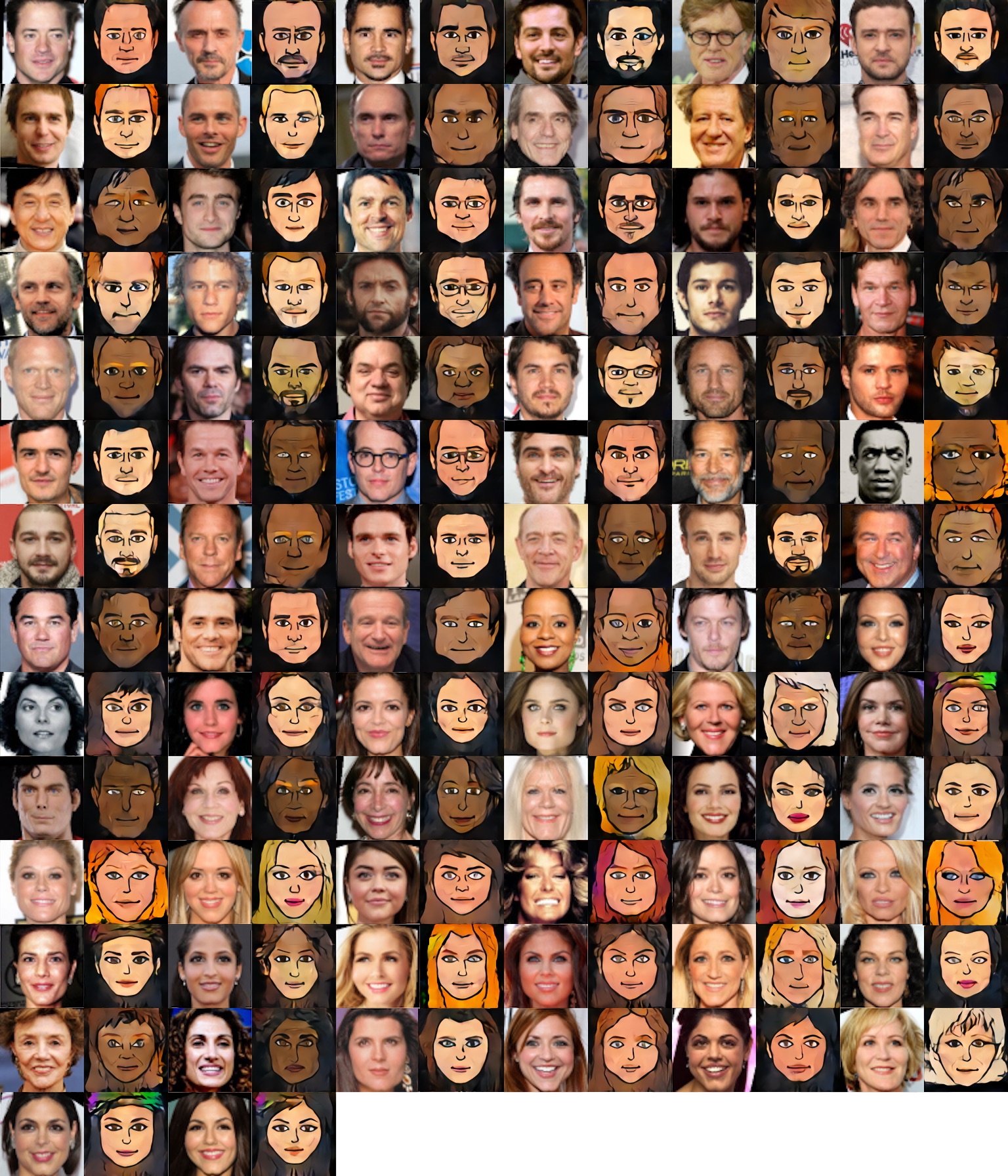}
\caption{\label{fig:allfacescrub}All 80 identities of the Facescrub dataset. The even columns show the results obtained for the images in the odd column to the left. Best viewed in color and zoom.}
\end{figure}

In Fig.~\ref{fig:allfacescrub} we show the full set of identities of the Facescrub dataset, and their corresponding generated emoji.

\newpage
\section{The effect of super-resolution}
\label{sec:superres}
\begin{figure}[H]
\centering
\begin{tabular}{cccc}
\includegraphics[trim=0 0 0 0, clip, width=0.230235\linewidth]{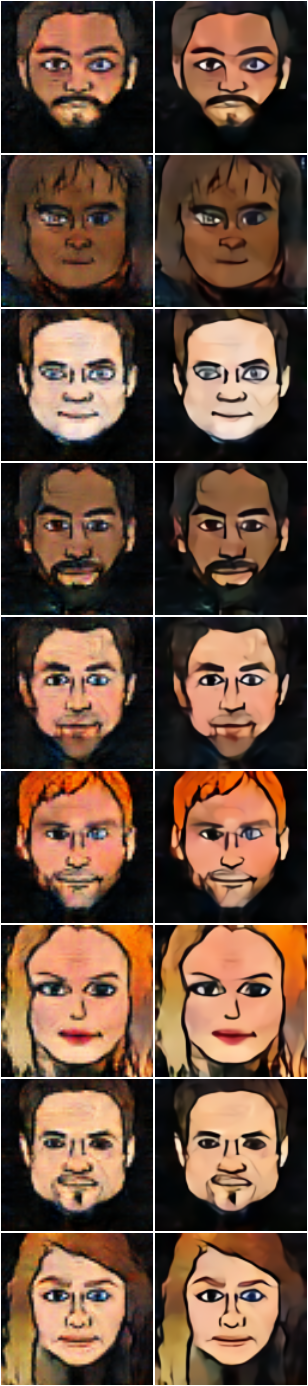}&
\includegraphics[trim=0 0 0 0, clip, width=0.230235\linewidth]{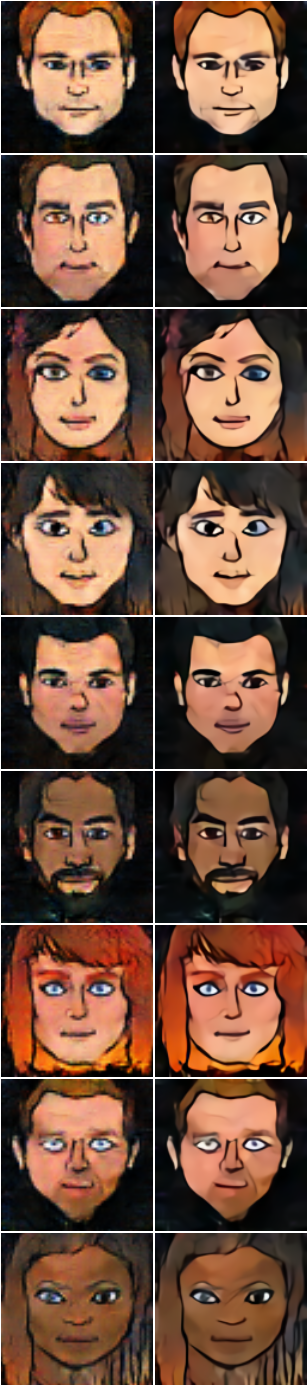}&
\includegraphics[trim=0 0 0 0, clip, width=0.230235\linewidth]{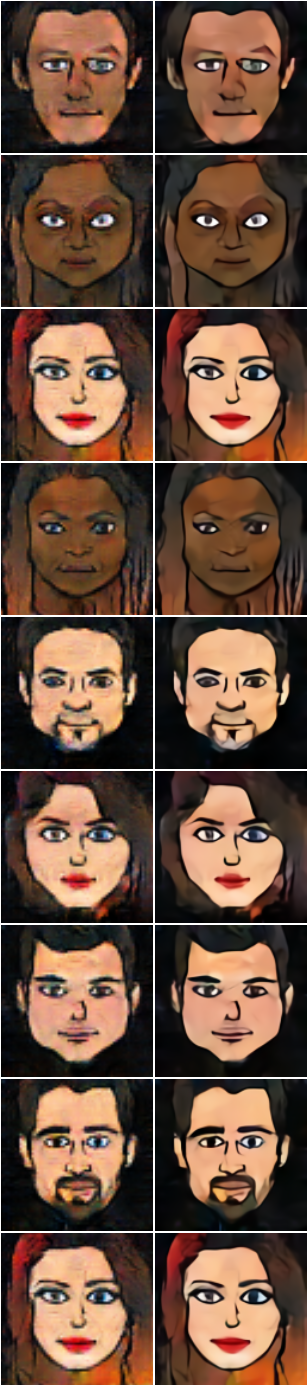}&
\includegraphics[trim=0 0 0 0, clip, width=0.230235\linewidth]{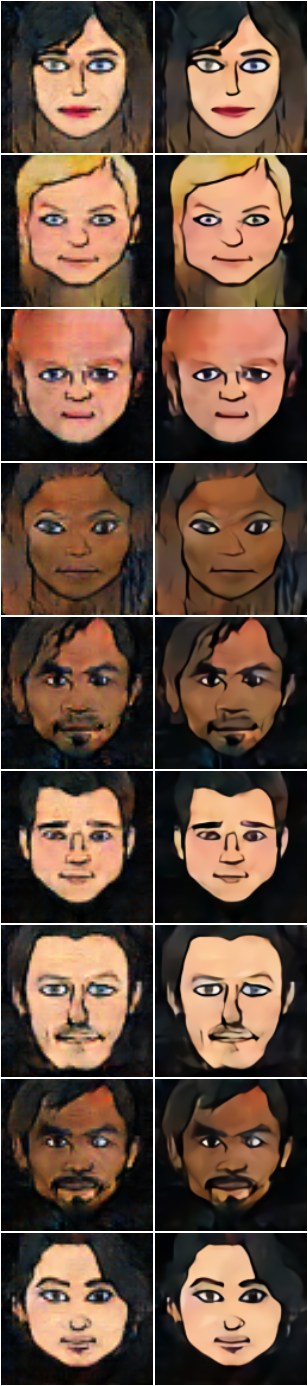}
\end{tabular}
\caption{\label{fig:superres} The images in Fig.~\ref{fig:IBtagB} above with (right version) and without (left version) applying super-resolution. Best viewed on screen.}
\end{figure}

As mentioned in Sec.~\ref{sec:expa}, in order to upscale the $64 \times 64$ output to print quality, the method of ~\cite{superres} is used. Fig.~\ref{fig:superres} shows the effect of applying this postprocessing step.
\section{The basis elements of the face representation}
\label{sec:e10}
\begin{figure}[H]
\centering
\includegraphics[trim=0 335 0 0, clip, width=.99\linewidth]{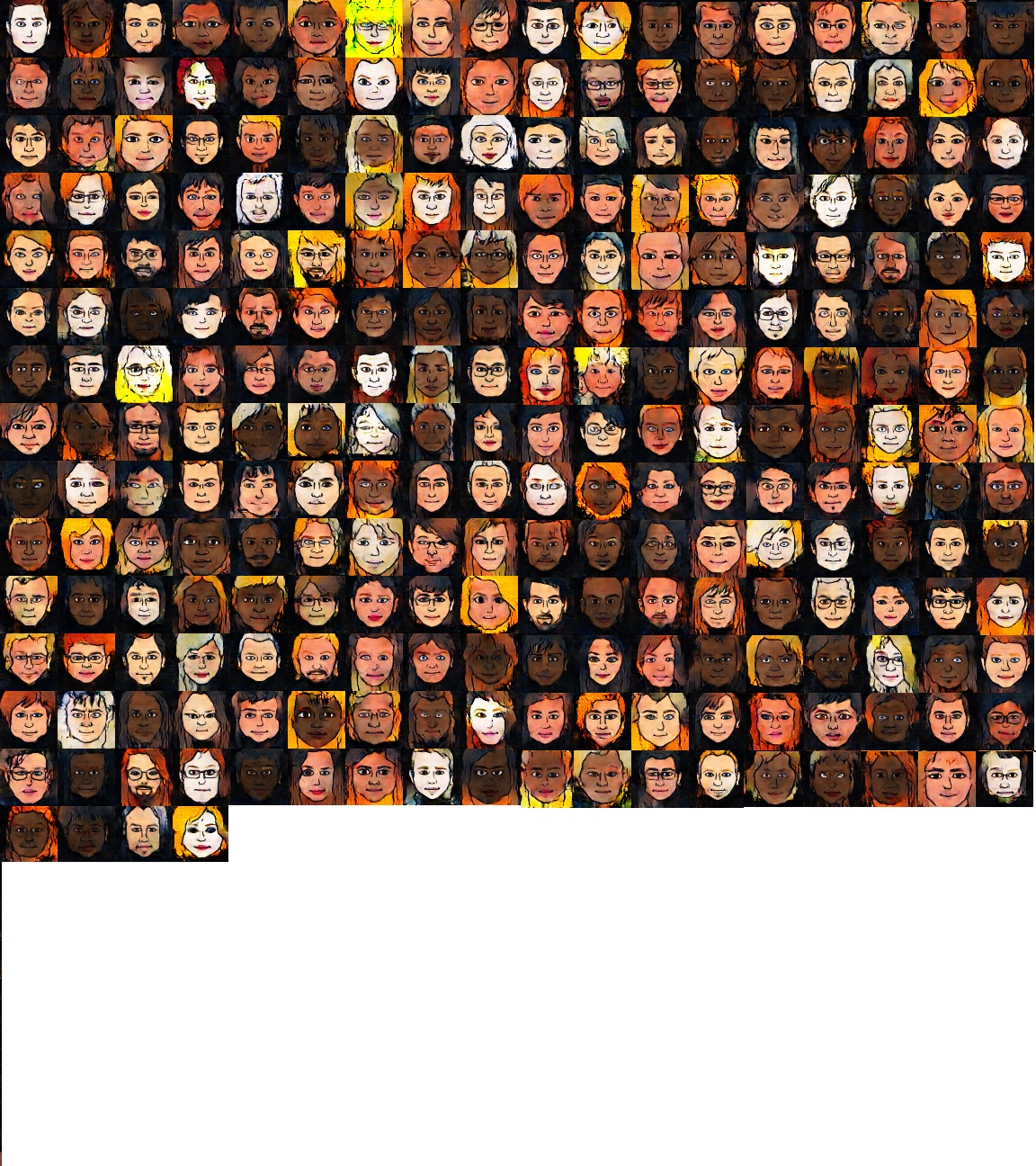}
\caption{\label{fig:e10} The emoji visualization of the standard basis vectors in the space of the face representation, i.e., $g(e_1)$,...,$g(e_{256})$, where $e_i$ is the $i$ standard basis vector in $\mathbb{R}^{256}$.}
\end{figure}

Fig.~\ref{fig:e10} depicts the face emoji generated by $g$ for the standard basis of the face representation~\citep{deepface}, viewed as the vector space $\mathbb{R}^{256}$. 
\section{Domain transfer in the reverse direction}
\label{sec:humanizer}

\begin{figure}[H]
\centering
\subfloat[]{
\includegraphics[trim=0 0 0 0, clip, height=0.5246157\linewidth]{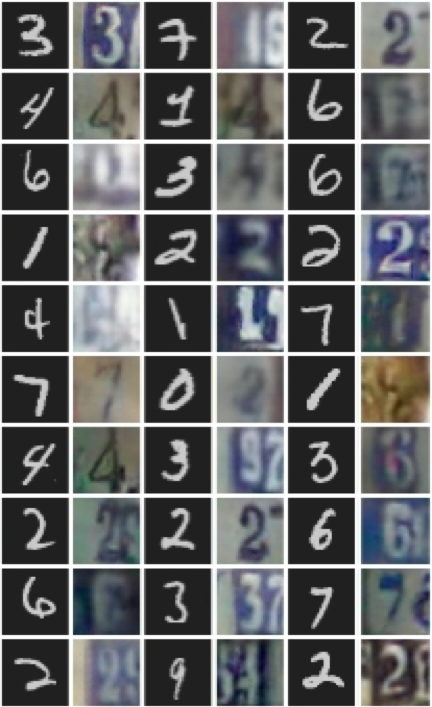}
}
\subfloat[]{
\includegraphics[trim=0 20 0 20, clip, height=0.5246157\linewidth]{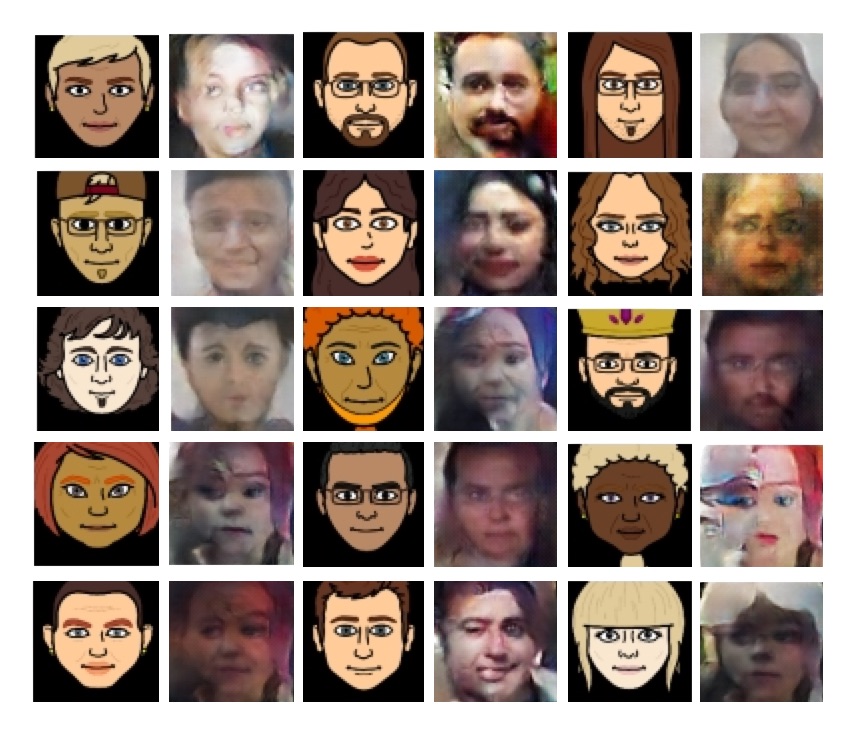}
}
\caption{\label{fig:rev}Domain transfer in the other direction (see limitations in Sec.~\ref{sec:conclusions}). Input (output) in odd (even) columns. (a) Transfer from MNIST to SVHN. (b) Transfer from emoji to face photos.}
\end{figure}

For completion, we present, in Fig.~\ref{fig:rev} results obtained by performing domain transfer using DTNs in the reverse direction of the one reported in Sec.~\ref{sec:expa}.


\vspace{-.2in}

\end{document}